%% file: emnlp2023.tex
\pdfoutput=1
\documentclass[11pt]{article}

\usepackage[]{EMNLP2023}

\usepackage{times}
\usepackage{latexsym}

\usepackage[T1]{fontenc}
\usepackage[utf8]{inputenc}

\usepackage{microtype}

\usepackage{inconsolata}
\usepackage{graphicx}
\usepackage{tabularx}
\usepackage{multicol}
\usepackage{multirow}
\usepackage{amsmath} % Required for the align environment
\usepackage{subfigure}
\usepackage{booktabs}
\usepackage{diagbox}
\usepackage[normalem]{ulem}
\usepackage{stfloats}
\useunder{\uline}{\ul}{}

\title{Time-Aware Representation Learning for Time-Sensitive Question Answering}

\author{Jungbin Son \\
  KAIST  \\
  \texttt{sonjbin@kaist.ac.kr} \\\And
  Alice Oh \\
  KAIST \\
  \texttt{alice.oh@kaist.edu} \\}

\begin{document}
\maketitle

\input{texts/0_abstract}

\section{Introduction}
\input{texts/1_introduction}

\section{Related Work}
\input{texts/2_relatedwork}

\section{Method}

\input{texts/3_method}

\section{Experimental Setup}
\input{texts/4_experiments}

\section{Result and Discussion}
\input{texts/5_results}

\section{Conclusion}
\input{texts/6_conclusion}

\section*{Ethical Consideration}
\input{texts/7_ethical}

\section*{Limitations}
\input{texts/8_limiation}

\bibliography{emnlp2023-latex/anthology_cus}
\bibliographystyle{acl_natbib}
\clearpage

\section*{Appendix}
\appendix

\input{texts/9_appendices}
\end{document}

%% file: texts/0_abstract.tex
\begin{abstract}

Time is one of the crucial factors in real-world question answering (QA) problems. However, language models have difficulty understanding the relationships between time specifiers, such as `after' and `before', and numbers, since existing QA datasets do not include sufficient time expressions. To address this issue, we propose a \textbf{T}ime-\textbf{C}ontext aware \textbf{Q}uestion \textbf{A}nswering (TCQA) framework. We suggest a \textbf{T}ime-\textbf{C}ontext dependent \textbf{S}pan \textbf{E}xtraction (TCSE) task, and build a time-context dependent data generation framework for model training. Moreover, we present a metric to evaluate the time awareness of the QA model using TCSE. The TCSE task consists of a question and four sentence candidates classified as correct or incorrect based on time and context. The model is trained to extract the answer span from the sentence that is both correct in time and context. The model trained with TCQA outperforms baseline models up to 8.5 of the F1-score in the TimeQA dataset.\footnote{Our dataset and code are available at \url{https://github.com/sonjbin/TCQA}}

\end{abstract}

% 
% In the TCSE task, the passage includes four types of sentences that depend on whether the questions and sentences are in line with respect to time and context.
% TCSE task includes four types of sentences, which are generated based on a pre-defined template, and the sentences are labeled correct or incorrect based on two aspects: time and context.
% The TCSE task consists of a question and four different answer candidates generated by a pre-defined template. Then, the candidates are classified as correct or incorrect based on time and context.

%% file: texts/1_introduction.tex
\begin{figure} [t]
  \centering
\includegraphics[width=1.0\linewidth]{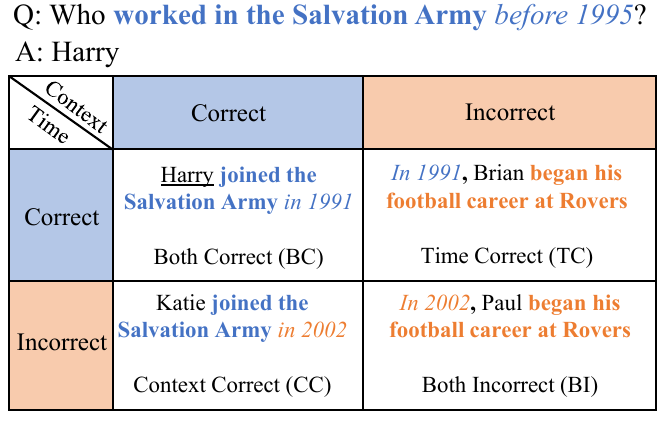}
\caption{Example case of Time-Context dependent Span Extraction (TCSE) task. The passage consists of four types of sentences that depend on whether the sentences match the time and context of the question. The target span is `Harry' in this example.}\label{fig:TCSE}
\end{figure}

Question Answering (QA) models \citep{devlin-etal-2019-bert, Clark2020ELECTRA:} have achieved significant success in recent years. However, most existing QA models fail to understand time \citep{chen2021a} since most QA datasets \citep{rajpurkar-etal-2018-know, kwiatkowski2019natural} lack temporal information. Ignoring temporal constraints when answering questions can lead to inaccurate or unreliable results \citep{CHEN2022109134}. For instance, as shown in Figure \ref{fig:TCSE}, neglecting the time while extracting the answer may lead to the selection of an incorrect entity, `Katie'.

To overcome this limitation, language models must be able to incorporate temporal information into their comprehension of the context in which a question is asked. This requires the model to recognize temporal expressions within the text and understand the relationship between the time specifiers and numerical values. For example, asking about anything that happened `after 2020' and `before 2020' are entirely different, even though they include the same number. Therefore, models must be capable of comprehending the connection between time specifiers and numbers beyond simple numerical comparisons.

This study aims to investigate methods for enhancing the performance of QA models in time-sensitive tasks. Specifically, we aim to develop a model that can process temporal information and utilize it to answer time-sensitive questions precisely. Injecting time awareness and numeracy into QA models is challenging since there are many possible temporal expressions, and the model must consider time information as an independent part of the context. Therefore, we propose a Time-Context aware Question Answering (TCQA) framework to achieve this issue. We train the model through Time-Context Dependent Span Extraction (TCSE) task and contrastive time representation learning.

In this paper, our contributions are:
\begin{itemize}
    \item We propose a TCQA framework that involves TCSE and contrastive time representation learning, and generate synthetic data to enhance temporal reasoning ability to understand time expressions. 
    \item We demonstrate that training the model with TCQA can improve the time awareness of QA models.
    \item We introduce a new metric to evaluate QA models in terms of time and context awareness.
\end{itemize}

% We will release the synthetic dataset and code to facilitate further research in time-sensitive QA.

%% file: texts/2_relatedwork.tex
Several previous works have addressed the issue of temporal reasoning in question answering using knowledge graphs. \citet{Shang2022ImprovingTS} proposed a novel framework for handling complex temporal questions that involve time ordering. \citet{saxena2021question} jointly train the model using text with timestamps. However, these approaches may not be sufficient for time-sensitive QA tasks, as temporal knowledge graphs typically handle only structured time information such as (Barack Obama, position held, President of USA, [2009, 2017]).

Despite these efforts, there remains a gap in research regarding handling various time expressions and numerical reasoning in time-sensitive QA tasks. \citet{chen2021a} attempted to address this gap by constructing a dataset containing time-sensitive question-passage pairs. Their analysis revealed that existing language models often fail to adequately consider temporal constraints in such tasks, resulting in significantly lower performance than humans.

%% file: texts/3_method.tex
We present an approach to improve the performance of models in time-sensitive QA tasks by proposing a Time-Context aware QA (TCQA) framework. 
\begin{figure*} [t]
  \centering
\includegraphics[width=1.0\linewidth]{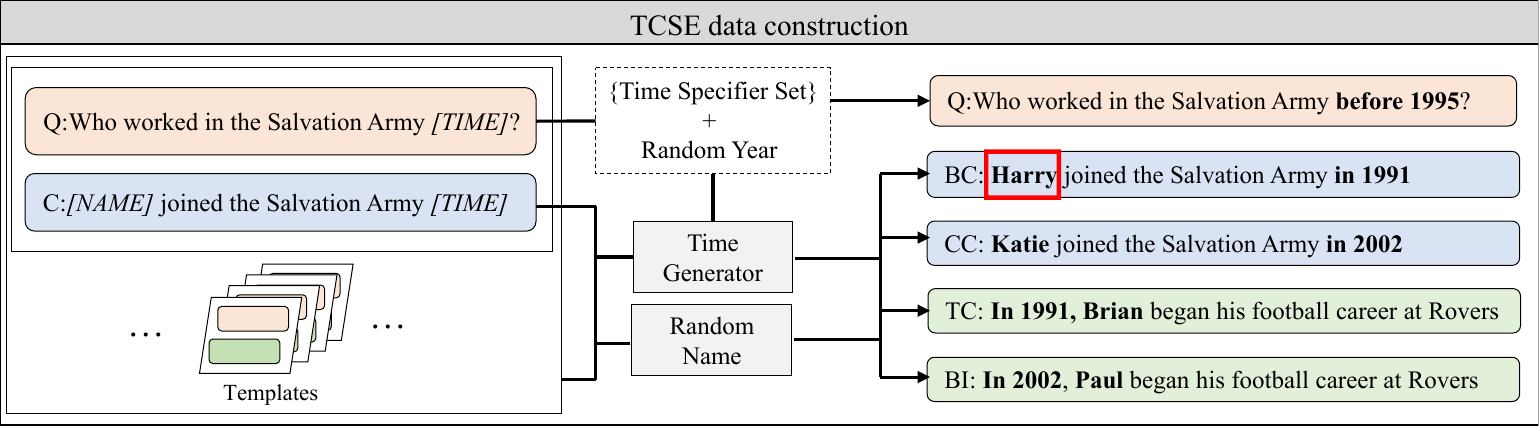}
\caption{Four types of candidates, namely BC, CC, TC, and BI, are derived from question-context templates via time expression generation and random name.}\label{fig:datac}
\end{figure*}

\subsection{Synthetic Time-Sensitive Data Generation}\label{sec:generation}

Data generation for TCSE involves constructing question-context templates. A question-context template is a pair of questions in which the time constraint is masked, and a context in which time information and target entity are masked, as shown in Figure \ref{fig:datac}. 

We extract time-related sentences from Wikipedia articles. Then, a question is generated for each extracted sentence by using the generation model \citep{raffel2020exploring}. We create a template of the question and sentence pair by replacing the person entity and time expression with special tokens, \texttt{`[NAME]' and `[TIME]'}, respectively.

To obtain time-sensitive question-context pairs, we utilize a time pair generation process and we employ the `names' Python module that randomly generates a person's name.

We generate random time pairs through rule-based matching of time specifiers and years. To simplify template generation, we assume that all events continue indefinitely when generating year numbers. We adopt seven time specifiers \{in, after, since, before, until, between, from\}. We generate positive time expressions that match the time range of the question and negative time expressions that does not. We exclusively use the time specifier `in' when generating time expressions for the context to facilitate model training. For example with rule-based matching, if the question time is `before 1995', then positive time is the year smaller than 1995, and negative time is the year greater than 1995. We randomly select one of the context templates to obtain a negative context.

As depicted in Figure \ref{fig:datac}, we get positive and negative context and time for each question. This allows us to produce sentences that are correct in both context and time (BC), only in context (CC), only in time (TC), and are incorrect in both (BI) for the corresponding question.

\subsection{Time-Context Aware Question Answering}

\subsubsection{Time-Context dependent Span Extraction} 
We train the model in a multi-task setting using both reading comprehension and TCSE tasks. The loss for the reading comprehension task, denoted as $L_{RC}$, is calculated by the sum of cross-entropy loss between ground truth and predicted distribution of start and end indices. Similarly, the TCSE task adopts the same loss function, but with the answer span set as the target entity in `BC' context.

\subsubsection{Contrastive Representation Learning}
In order to enhance the time-awareness, we employ contrastive learning. We construct contrastive samples with TCSE data by pairing questions and contexts. For each sample within TCSE data, only BC context over four contexts corresponds to the question. Therefore, one positive pair (BC) and three negative pairs (TC, CC, BI) are constructed for each TCSE data sample. 

Contrastive loss is calculated based on the cosine similarity between the embedding of question ($v_q$) and context ($v_c$) for each pair. It is desirable for the distance between positive pairs to be minimized, while the distance between negative pairs should be maximized. Consequently, the contrastive label, denoted as Y, is assigned: Y=1 signifies that the context corresponds to a positive sample of the question, whereas Y=0 indicates that the context represents a negative sample. Subsequently, the contrastive loss is computed as follows:
\begin{equation}
    s_{i} = CosineDistance(v_{q_i}, v_{c_i})
\end{equation}
\begin{equation}
    \begin{split}
    L_{Contrast} = \sum_i[ w_p Y*exp(s_i) \\
    +w_n(1-Y)*exp(1-s_i))]\end{split}
\end{equation}
$w_p$ and $w_n$ is the weight of the positive and negative sample, respectively.

\subsubsection{Joint Training}
The final loss is defined as a weighted sum of answer-span prediction loss ($L_{RC}$), TCSE loss ($L_{TCSE}$) and contrastive loss ($L_{Contrast}$):
\begin{equation}
    L_{total} = L_{RC}+\lambda_{T}*L_{TCSE}+\lambda_{C}*L_{Contrast}
\end{equation}

\subsection{Evaluation Metric of Time Awareness}
We propose a new evaluation metric for measuring the time awareness of the model leveraging TCSE. Since the TCSE dataset labels which sentence is correct in terms of the time or the context, it is possible to determine whether the model extracted the answer from the correct time or context. Specifically, if the model correctly extracts the answer from BC or TC sentence, it indicates that the model finds the answer in the correct time range. Similarly, if the model extracts the answer from BC or CC sentences, it indicates that the model identified the correct context. Therefore, the Time Awareness (TA) and the Context Awareness (CA) scores are calculated by the ratio of cases in which the model extracts the answer in the correct time range or context, respectively. Awareness scores are calculated with the following equations:  
\begin{align}
\text{TA}
= \frac{|BC|+|TC|}{\text{(\# \ of \ questions)}}\nonumber \\
\text{CA}
= \frac{|BC|+|CC|}{\text{(\# \ of \ questions)}}\nonumber \\
\label{eq:TACA}
\end{align}
Where |BC|, |TC|, |CC| indicate the number of questions that the model extracts the answer in BC, TC, CC, respectively.
Then, Time-Context awareness score (TC-score) is calculated as the harmonic mean of TA and CA:
\begin{equation}
\text{TC-score}
= 2\times  \frac{TA \times CA}{TA+CA} 
\label{eq:TCAS}
\end{equation}

TC-score allows for a comprehensive evaluation of a model's performance in terms of both time and context awareness.

%% file: texts/4_experiments.tex
\subsection{Dataset}
\textbf{TimeQA} \cite{chen2021a} is a reading comprehension dataset that involves complex temporal reasoning. TimeQA consists of two subsets, easy and hard-mode, which differ in the level of difficulty of temporal reasoning required. We use a hard-mode dataset as it involves reasoning with more complex time expressions.
% , such as matching two time ranges. The resolution of questions in the hard-mode dataset is not attainable through text-matching only.

To evaluate the TC-score of the model, we generate a test set of TCSE task using time-related sentences from Wikipedia pages not included in the training data.

As a result, we generated 10,302 templates, and we generated 118,104 TCSE data for training, and 9323 TCSE data for tests from templates.

\subsection{Baselines}
\textbf{BERT} \cite{devlin-etal-2019-bert}, \textbf{RoBERTa} \cite{liu2019roberta} and \textbf{ALBERT} \cite{conf/iclr/LanCGGSS20} is a large pre-trained language model largely used in QA tasks. In our experiments, we use the base model fine-tuned with SQuAD2.0 \cite{rajpurkar-etal-2018-know}. \\
\textbf{BigBird} \citep{zaheer2020big} is a language model that was developed to handle long sequence input.
In our experiments, we use the RoBERTa \cite{liu2019roberta} base BigBird model fine-tuned with Natural Questions (NQ) \cite{kwiatkowski2019natural}.

%% file: texts/5_results.tex
\begin{table*}[ht!]
\centering
\begin{tabular}{@{}c|cc|cc|cc|cc@{}}
\toprule
Model               & \multicolumn{2}{c|}{$BERT_{base}$} & \multicolumn{2}{c|}{$RoBERTa_{base}$} & \multicolumn{2}{c|}{$ALBERT_{base}$} & \multicolumn{2}{c}{$BigBird_{RoBERTa}$} \\ \midrule
Metric               & EM          & F1       & EM           & F1       & EM          & F1       & EM           & F1   \\ \midrule
Baseline       & 19.95        & 26.25    &29.89        &38.5         &24.66         &34.5      & 44.43        & 53.21        \\ \midrule
\begin{tabular}[c]{@{}c@{}}+TCQA\end{tabular} & \textbf{25.63}       & \textbf{34.75}  &\textbf{30.86}         &\textbf{39.03}     &\textbf{27.36}         &\textbf{35.48}     & \textbf{46.31}        & \textbf{54.26}        \\ \bottomrule
\end{tabular}
\caption{Performance of baseline models, model trained with timeQA data, and model trained with the proposed method. We evaluate the model on the TimeQA test dataset; three runs average all results. Our method outperforms the baseline model.}
\label{tab:TimeQA}
\end{table*}

% We present the experimental result in this section. We show that the model trained with TCQA outperforms baseline models in a time-sensitive QA task. We also demonstrate that TCSE data can be employed to assess the time and context awareness of QA models.

\subsection{Time-Sensitive Question Answering}
We evaluate time-sensitive QA performance on TimeQA \cite{chen2021a} dataset. We show the result in Table \ref{tab:TimeQA}, demonstrating that training the model with TCQA outperforms the baseline models. BERT model further trained on TCQA shows a significant performance improvement of 8.5 F1-score compared to the model trained only on TimeQA. This improvement is the result of the model learning to distinguish correct time expressions. The performance gap between BigBird and others can be attributed to their maximum input length difference.

 \begin{table}[t]
\centering
\begin{tabular}{@{}c|c|ccc@{}}
\toprule
   FT dataset             & F1    &TA   &CA   &TC-score       \\ \midrule
 NQ            & 35.92    &51.48     &\textbf{88.78}     &65.16   \\ 
TimeQA               & \textbf{53.56}    &\textbf{67.96}     &79.32     &\textbf{73.21}  \\ \bottomrule
\end{tabular}
\caption{Comparison among the F1-score in TimeQA, and score in TCSE task: Time Awareness (TA), Context Awareness (CA), and Time-Context awareness score (TC-score) of $BigBird_{RoBERTa}$ model according to the data used for fine-tuning.}
\label{tab:TCAS}
\end{table}

\subsection{Time and Context Awareness}
We evaluate the model's time awareness and context awareness using the TC-score. Table \ref{tab:TCAS} indicates that the F1-score and TA exhibit similar trends, implying that TA is a reliable indicator of time awareness. We observed that training with TimeQA resulted in a decrease in contextual understanding, as evidenced by an 9.46-point drop in CA. The results suggest the importance of learning time expressions while maintaining contextual understanding. We utilized the TC-score to provide an overall assessment of the model's performance. We found that the model's contextual awareness decreased, but its time awareness improved significantly, resulting in improved TC-score. We do not perform TC-score on models trained with TCQA, because the model has already learned the TCSE task. Alternatively, TCQA is assessed using an alternative approach in Appendix \ref{app:RC}.

\subsection{Analysis on Time Specifier}
\begin{figure}[h]
    \centering
    \includegraphics[width=0.8\columnwidth]{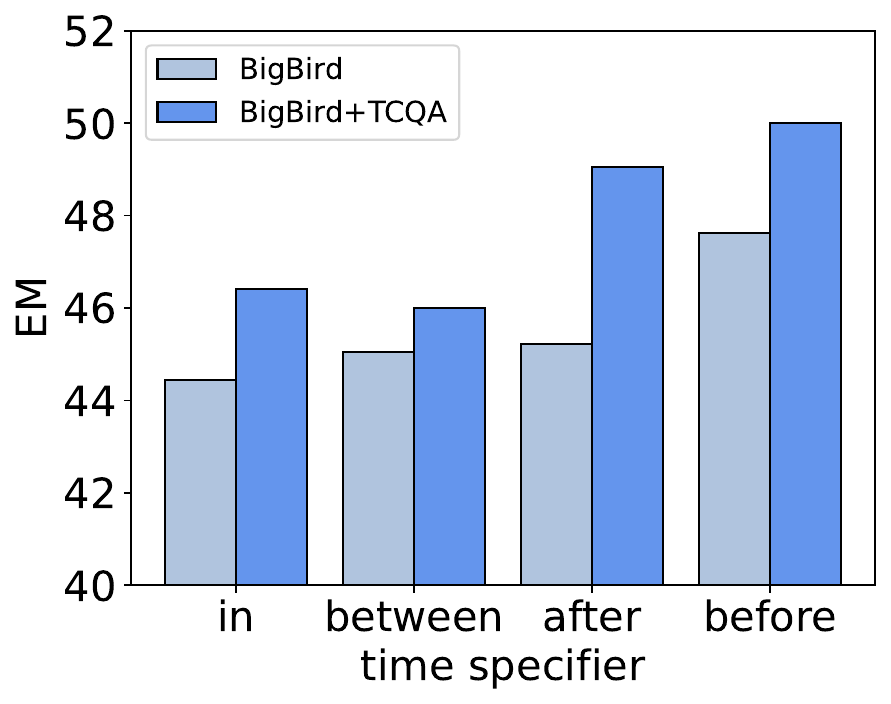}
    % \subfigure[\centering EM]{\includegraphics[width=0.8\columnwidth]{figures/specifier_EM.pdf}}
    % \\
    % \subfigure[\centering F1]{\includegraphics[width=0.8\columnwidth]{figures/specifier_F1.pdf}}

    \caption{EM score on TimeQA of Bigbird according to the time specifier included in the question.}
    \label{fig:specifier}
\end{figure}

We analyze model performance on TimeQA according to the time specifier included in the question. Figure \ref{fig:specifier} shows the EM score difference for four kinds of time specifiers: \{in, between, after, before\}. There are comparatively substantial improvements in model performance on time specifiers `after' and `before'. This improvement demonstrates that TCQA effectively trains the model to understand the time range. However, the performance improvements on time specifier `between' is comparatively low since it is more difficult as it requires simultaneous consideration of two distinct time ranges.

\begin{table*}[ht!]
\centering
\begin{tabular}{@{}c|cc|cc|cc|cc@{}}
\toprule
Model               & \multicolumn{2}{c|}{$BERT_{base}$} & \multicolumn{2}{c|}{$RoBERTa_{base}$} & \multicolumn{2}{c|}{$ALBERT_{base}$} & \multicolumn{2}{c}{$BigBird_{RoBERTa}$} \\ \midrule
Metric               & EM          & F1       & EM           & F1       & EM          & F1       & EM           & F1   \\ \midrule
\begin{tabular}[c]{@{}c@{}}w/o CRL\end{tabular}  & 23.1        & 31.99    &30.7        &38.16         &25.83         &34.65      & \textbf{46.7}        & \textbf{54.44 }        \\ \midrule
w/ CRL   & \textbf{25.63}       & \textbf{34.75}  &\textbf{30.86}         &\textbf{39.03}     &\textbf{27.36}         &\textbf{35.48}     & 46.31        & 54.26          \\ \bottomrule
\end{tabular}
\caption{Results of the ablation study of the contrastive learning}
\label{tab:ablation}
\end{table*}

\subsection{Ablation Study}
An ablation study was conducted to assess the impact of contrastive representation learning. We compare the performance of the model trained with and without adding contrastive loss in Table \ref{tab:ablation}. Adding contrastive loss resulted in an improvement in the model performance of BERT, RoBERa and ALBERT model. However, it led to a slight decrease in performance for the Bigbird model. The reason for this discrepancy of result is that the vector embedding size of the Big Bird model is eight times greater than that of the other models. Consequently, we can infer that the data used for contrastive learning was insufficient.

\subsection{Reading Comprehension Performance}\label{app:RC}
We conduct a comparative analysis of models on the SQuAD v2 dataset to investigate the effect of TCQA on context awareness. Table \ref{tab:squad} demonstrate that TCQA mitigates the degradation in context awareness resulting from fine-tuning on TimeQA. This improvement caused by training the model using time-context dependent data.

\begin{table}[h]
\centering
\begin{tabular}{@{}c|cc@{}}
\toprule

               & EM            & F1        \\ \midrule
FT on TimeQA        & 34.74          & 46.49        \\
+ TCQA         & \textbf{36.12}         &\textbf{48.1}     \\\bottomrule
\end{tabular}
\caption{ Performance of BigBird model on SQuAD v2 development set.}
\label{tab:squad}
\end{table}

%% file: texts/6_conclusion.tex
In this paper, we demonstrated that existing QA models are inadequate in understanding time expressions. To address this problem, we proposed TCQA, which enables models to learn time expressions while maintaining their understanding of context. We constructed question-context templates to generate time-context dependent data for TCSE and contrastive learning, and jointly trained the model. Our experimental results showed that TCQA improves the performance of QA models on TimeQA. Additionally, we proposed a new evaluation metric, TC-score, and showed a gap in performance between models in terms of time and contextual understanding. Future research should focus on advancing temporal reasoning capabilities beyond the comprehension of simple temporal expressions.

%% file: texts/7_ethical.tex
This paper presents a synthetic data generation framework that modifies time information and name while retaining the original text. Notably, this approach does not produce any unintended harmful effects, as it does not alter the semantic content of the original text beyond the specified modifications.

%% file: texts/8_limiation.tex
Limitation of our approach is that TCSE does not cover all kinds of time expressions because we construct the data with only seven time specifiers. Although it is possible to enhance the model's time awareness by adding additional time expressions, our experiments showed that the inclusion of only these seven led to a performance improvement.

%% file: texts/9_appendices.tex
% \subsection{Template Construction for TCSE}
% Fig \ref{fig:dataconstruction} shows the process of question-context templates. Time-related sentences are extracted from Wikipedia, comprising seven time specifiers, {in, after, since, before, until, between, from}, along with year information. A regular expression (regex) is employed to identify sentences containing only one time specifier with year. Even though this method could not detect all time expressions in the text, it extracts sufficient time-related sentences.

% \begin{figure*}
%     \centering
%     \includegraphics[width=1.0\textwidth]{figures/DataConstruction.pdf}

%     \caption{TCSE data construction from generated templates }
%     \label{fig:dataconstruction}
% \end{figure*}

\section{Hyper Parameter Setting}
\subsection{Analysis on $\lambda_T$ and $\lambda_C$}
\begin{figure}[h]
    \centering
    \subfigure[\centering EM]{\includegraphics[width=0.49\columnwidth]{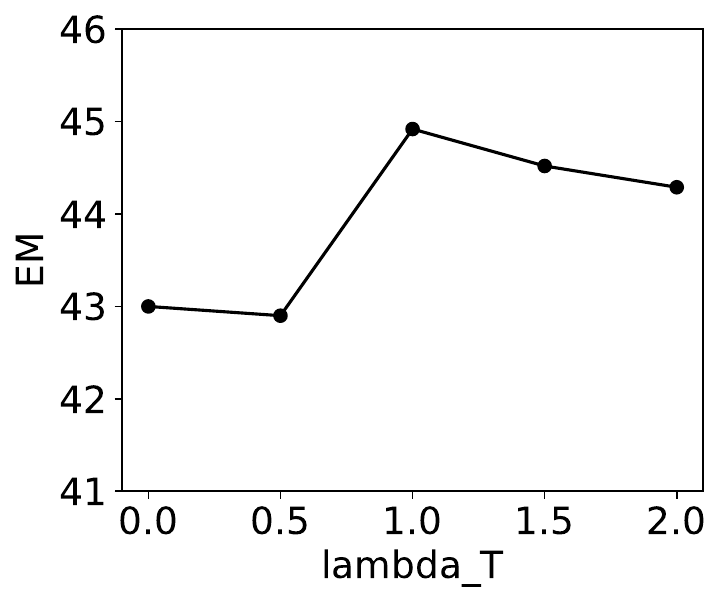}}
    \subfigure[\centering F1]{\includegraphics[width=0.49\columnwidth]{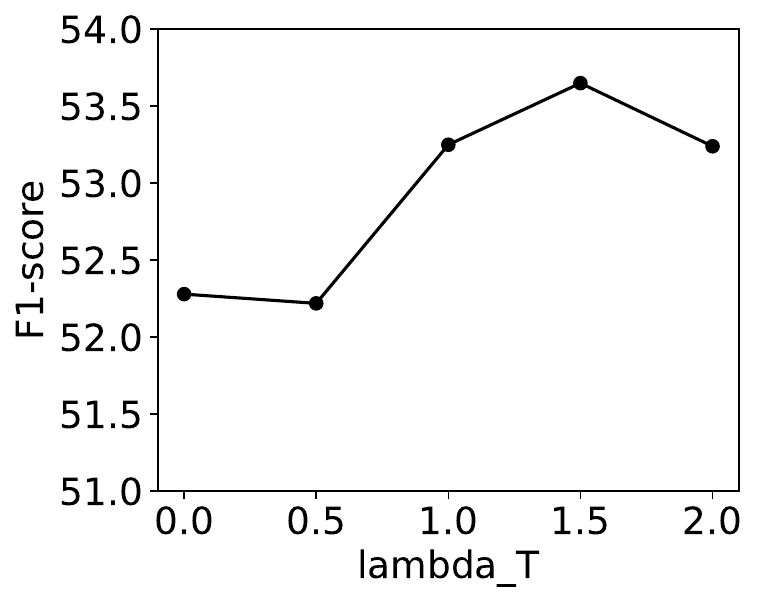}}

    \caption{Analysis on $\lambda_T$ for time-sensitive question answering for TimeQA dataset with $Bigbird_{RoBERTa}$ model. We increase lambda from 0 to 2.0: \{0, 0.5, 1.0, 1.5, 2.0\}. Increasing lambda improves time-sensitive question answering performance until $\lambda=1.0$ and then decreases.}
    \label{fig:lambda_T}
\end{figure}

We observe the changes in model performance according to the value of $\lambda_T$. Figure \ref{fig:lambda_T} shows that the EM and F1-score increases with an increase in $\lambda_T$ until it reaches a value of 1.0 and 1.5, respectively. However, the model performance decreases when $\lambda_T$ was set to a value greater than  due to overfitting the TCSE task.

Additionally, we conduct an analysis of the changes in model performance as determined by the value of $\lambda_C$ while maintaining a $\lambda_T$ value fixed to 1.0. Model performance tends to decrease for $\lambda_C$ values greater than 0.5.

\begin{figure}[hbt!]
    \centering
    \subfigure[\centering EM]{\includegraphics[width=0.49\columnwidth]{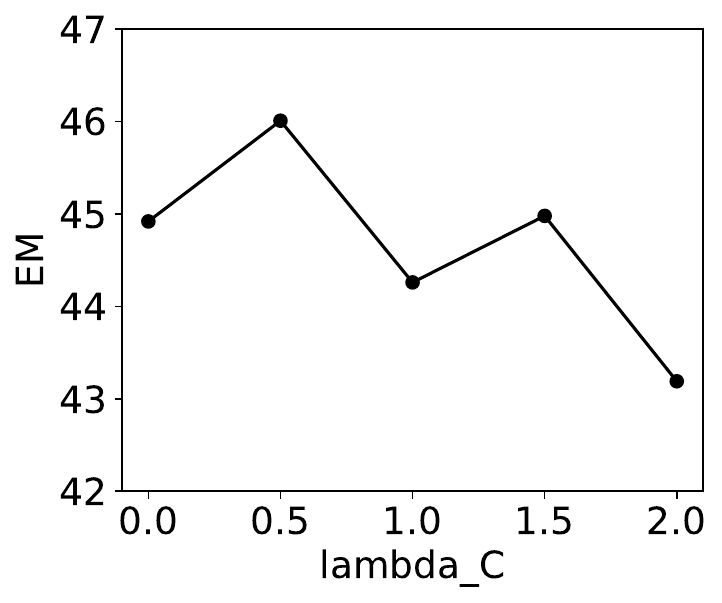}}
    \subfigure[\centering F1]{\includegraphics[width=0.49\columnwidth]{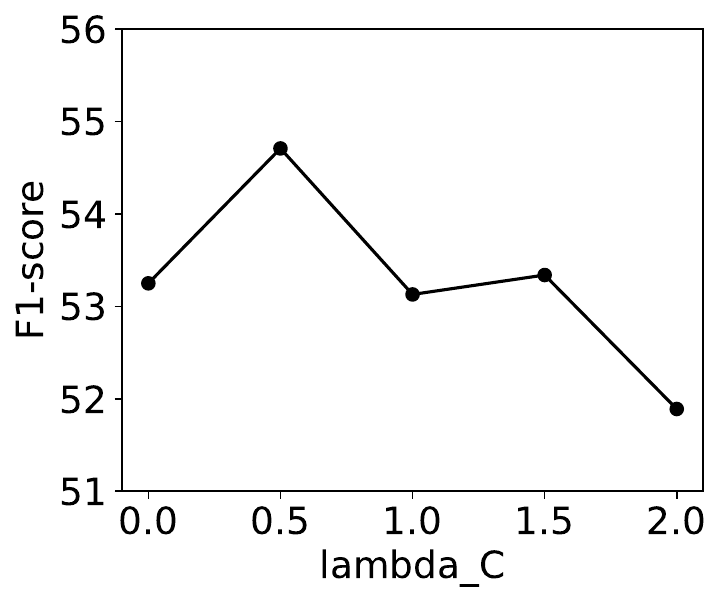}}

    \caption{Analysis on $\lambda_C$ for time-sensitive question answering for TimeQA dataset with $Bigbird_{RoBERTa}$ model. We increase lambda from 0 to 2.0: \{0, 0.5, 1.0, 1.5, 2.0\}.}
    \label{fig:lambda_C}
\end{figure}

\subsection{Effect of TCSE Dataset Size}

\begin{table}[ht]
\centering
\begin{tabular}{@{}c|cc@{}}
\toprule
                        & EM    & F1    \\ \midrule
$Bigbird_{RoBERTa}$          & 44.43 & 53.21 \\
+$TCSE_{1}$             &45.31  &53.92  \\
+$TCSE_{2}$             &\textbf{46.7}   &\textbf{54.44}   \\
+$TCSE_{4}$           &45.96  &54.26  \\ \bottomrule
\end{tabular}
\caption{Effect of TCSE according to the ratio of dataset size. $TCSE_{k}$ denotes that it employs TCSE data corresponding to k times the number of TimeQA dataset.}
\label{tab:dataszie}
\end{table}

To investigate the effect of the dataset size of TCSE on the model performance, we observe changes in performance according to the number of TCSE data. As shown in Table \ref{tab:dataszie}, utilizing a larger TCSE data than that of TimeQA yields a more substantial improvement on TimeQA until using two times the number of TimeQA dataset.

\section{Qualitative Analysis}
\subsection{Case Study}
To clearly understand our model's improvement in time awareness, we present a case study on the TimeQA dataset in Table \ref{tab:case}. Our model successfully finds the correct answer in the context with the correct time range. The model correctly answered a challenging question that required verifying whether the time range `between 1831 and 1833' matches with `from 1829 to 1835'. Furthermore, our model recognizes that a sentence containing the correct context but with an incorrect time range does not yield an answer.

\subsection{Example of TCSE Data}
We present examples of TCSE data to substantiate the reliability of synthetic data. Table \ref{tab:example} demonstrates the effectively generated questions and their corresponding sentences. Every sentence, `BC', `TC', `CC', `BI' , for each question are effectively generated and appropriately labeled in accordance with temporal and contextual considerations.

\begin{table*}[ht!]
\centering
\begin{tabularx}{1.0\textwidth}{p{3cm}Xp{2cm}p{2cm}}
\toprule
Question & Passage & \begin{tabular}[c]{@{}c@{}}Baseline\end{tabular} & \begin{tabular}[c]{@{}c@{}}Baseline\\ +TCQA\end{tabular} \\
\midrule
A: What position did John Pope take \textbf{between Sep 1831 and Nov 1833}? 
& ...
He served as a \textcolor{red}{member of the Kentucky Senate} \textit{from 1825 to 1829} , and ...

...
\textbf{From 1829 to 1835} , he served as the \textcolor{blue}{Governor of Arkansas Territory} .
...
 & \textcolor{red}{member of the Kentucky Senate} & \textcolor{blue}{Governor of Arkansas Territory}\\ \\

B: Sarah Bond was an employee for whom \textbf{in Feb 2011}?
 & ...
Bond was appointed Assistant Professor of Classics at the University of Iowa in 2014 , after holding an assistant professorship in Ancient and Early Medieval History at \textcolor{red}{Marquette University} \textit{from 2012} .
...
& \textcolor{red}{Marquette University} & \textcolor{blue}{Unanswerable}\\
\bottomrule
\end{tabularx}
\caption{A case study on TimeQA dataset: proposed model successfully (A) extracts the answer with the correct time and context and (B) detects an unanswerable question. }
\label{tab:case}
\end{table*}

\begin{table*}[ht!]
\centering
\begin{tabularx}{1.0\textwidth}{p{3cm}Xp{2cm}}
\toprule
Question & Example \\
\midrule
Who was General Counsel \textcolor{blue}{in 1481}? 

& \textbf{BC}: ... \textcolor{blue}{in 1473} ... Mitchell became General Counsel and a managing director.

\textbf{TC}: ... Abendroth, who was banned from working as a legal trainee \textcolor{blue}{in 1473}...

\textbf{CC}: ... \textcolor{red}{in 1483} ... Kimberly became General Counsel and a managing director.

\textbf{BI}: ... Jefferson Rash also attended the Bilderberg conference \textcolor{red}{in 1483} in St.
 \\ \\

Who played Asian Cup finals \textcolor{blue}{between 1566 and 1569}?

& \textbf{BC:} \textcolor{blue}{In 1567}, Julio participated with the team in the finals of the Asian Cup...

\textbf{TC:} Jon Kyl, who had represented the district \textcolor{blue}{in 1567}, said... 

\textbf{CC:} \textcolor{red}{In 1578}, Betty participated with the team in the finals of the Asian Cup... 

\textbf{BI:} ... golf womens Izod products were put on hiatus \textcolor{red}{in 1578} , but...

\\ \\ 
Who was loaned to Wolves \textcolor{blue}{before 2013}?

& \textbf{BC}: Matthew had a second loan spell at Wolves, as well as ... \textcolor{blue}{in 2007}...

\textbf{TC}: Rita went through...the last time an NFL team had done that was \textcolor{blue}{in 2007} ... 

\textbf{CC}: Martin had a second loan spell at Wolves, as well as ... \textcolor{red}{in 2020}... 

\textbf{BI}: Bell also played in a 3–1 defeat and a draw with West Germany \textcolor{red}{in 2020}...

\\ \\
Who was governor of ohio  \textcolor{blue}{from 2003 to 2004}?

& \textbf{BC}: David did not return ... until he served as Governor of Ohio \textcolor{blue}{in 2004} ...

\textbf{TC}: He also played \textcolor{blue}{in 2004}  AFC Asian Cup , as well as ...

\textbf{CC}: Sandra did not return... until he served as Governor of Ohio \textcolor{red}{in 2011} ... 

\textbf{BI}: Vecsei ... project designer on the construction of ... \textcolor{red}{in 2011}   ...
\\

\bottomrule
\end{tabularx}
\caption{Example of TCSE data for each time specifiers: in, between, before, from}
\label{tab:example}
\end{table*}

\section{Handling Long Sequence Input}
Since TimeQA \cite{chen2021a} contains long passages of more than 10,000 tokens, we split them into length intervals that correspond to the maximum input length of the models. During training, we use the context span that contains the indices of the answer span for answerable questions and only the first context span for unanswerable questions. We select the final answer as the maximum logit value among each split context during inference.

\section{Implementation Details}

We followed the implementation detail of TimeQA\footnote{\url{https://github.com/wenhuchen/Time-Sensitive-QA.git}} to train models using the TimeQA dataset. Baseline models are trained using Quadro RTX A6000 48GB, with a training batch size of 4, and a learning rate of 2e-5. Model fine-tuning per epoch took approximately 5 hours for BERT\footnote{\url{https://huggingface.co/bert-base-uncased}} and 12 hours for BigBird\footnote{\url{https://huggingface.co/vasudevgupta/bigbird-roberta-natural-questions}}.